\documentclass[conference, a4paper]{IEEEtran}
\ifCLASSINFOpdf
\else
\fi
\usepackage{graphicx}
\usepackage{subfigure}
\usepackage{float}
\usepackage{booktabs}
\usepackage{multirow}
\usepackage{multicol}
\usepackage{amsmath}
\usepackage{subfigure}
\usepackage{array}
\usepackage{cases}

\begin{document}
%
% paper title
% Titles are generally capitalized except for words such as a, an, and, as,
% at, but, by, for, in, nor, of, on, or, the, to and up, which are usually
% not capitalized unless they are the first or last word of the title.
% Linebreaks \\ can be used within to get better formatting as desired.
% Do not put math or special symbols in the title.
\title{Local Multiple Directional Pattern of Palmprint Image}

% author names and affiliations
% use a multiple column layout for up to three different
% affiliations

%\author{\IEEEauthorblockN{}
%\IEEEauthorblockA{\\}
%}

\author{\IEEEauthorblockN{Lunke Fei, Jie Wen, Zheng Zhang, Ke Yan, Zuofeng Zhong}
\IEEEauthorblockA{Shenzhen Graduate School, Harbin Institute of Technology, Shenzhen, China\\
Email: flksxm@126.com, wenjie@hrbeu.edu.cn, darrenzz219@gmail.com, yanke401@163.com, zfzhong2010@gmail.com}
}

% conference papers do not typically use \thanks and this command
% is locked out in conference mode. If really needed, such as for
% the acknowledgment of grants, issue a \IEEEoverridecommandlockouts
% after \documentclass

% for over three affiliations, or if they all won't fit within the width
% of the page, use this alternative format:
%
%\author{\IEEEauthorblockN{Michael Shell\IEEEauthorrefmark{1},
%Homer Simpson\IEEEauthorrefmark{2},
%James Kirk\IEEEauthorrefmark{3},
%Montgomery Scott\IEEEauthorrefmark{3} and
%Eldon Tyrell\IEEEauthorrefmark{4}}
%\IEEEauthorblockA{\IEEEauthorrefmark{1}School of Electrical and Computer Engineering\\
%Georgia Institute of Technology,
%Atlanta, Georgia 30332--0250\\ Email: see http://www.michaelshell.org/contact.html}
%\IEEEauthorblockA{\IEEEauthorrefmark{2}Twentieth Century Fox, Springfield, USA\\
%Email: homer@thesimpsons.com}
%\IEEEauthorblockA{\IEEEauthorrefmark{3}Starfleet Academy, San Francisco, California 96678-2391\\
%Telephone: (800) 555--1212, Fax: (888) 555--1212}
%\IEEEauthorblockA{\IEEEauthorrefmark{4}Tyrell Inc., 123 Replicant Street, Los Angeles, California 90210--4321}}

% use for special paper notices
%\IEEEspecialpapernotice{(Invited Paper)}

% make the title area
\maketitle

% As a general rule, do not put math, special symbols or citations
% in the abstract
\begin{abstract}
Lines are the most essential and discriminative features of palmprint images, which motivate researches to propose various line direction based methods for palmprint recognition. Conventional methods usually capture the only one of the most dominant direction of palmprint images. However, a number of points in palmprint images have double or even more than two dominant directions because of a plenty of crossing lines of palmprint images. In this paper, we propose a local multiple directional pattern (LMDP) to effectively characterize the multiple direction features of palmprint images. LMDP can not only exactly denote the number and positions of dominant directions but also effectively reflect the confidence of each dominant direction. Then, a simple and effective coding scheme is designed to represent the LMDP and a block-wise LMDP descriptor is used as the feature space of palmprint images in palmprint recognition. Extensive experimental results demonstrate the superiority of the LMDP over the conventional powerful descriptors and the state-of-the-art direction based methods in palmprint recognition.
\end{abstract}

% no keywords

% For peer review papers, you can put extra information on the cover
% page as needed:
% \ifCLASSOPTIONpeerreview
% \begin{center} \bfseries EDICS Category: 3-BBND \end{center}
% \fi
%
% For peerreview papers, this IEEEtran command inserts a page break and
% creates the second title. It will be ignored for other modes.
\IEEEpeerreviewmaketitle

\section{Introduction}
% no \IEEEPARstart
As a relative new biometric, palmprint based recognition has great potential to achieve more reliable performance due to its rich highly discriminative features, including not only the special principal lines and wrinkles but also the unique ridge patterns and minutiae points \cite{b1}\cite{b2}\cite{b3}. Among them, the ridge and minutiae based features can only be detected from high-resolution palmprint image with more than 400 ppi, which are mainly used for forensic applications \cite{b4}. Due to the rapid development of e-commerce applications, such as the Apple Pay, the low-resolution palmprint image based recognition has recently received increasing attentions \cite{b5}. This paper also focuses on the low-resolution palmprint image analysis.

In a low-resolution palmprint image, lines are clearly visible and deemed to be the most significant features of palmprint images. Huang et al. \cite{b6} extracted the stable principle lines for palmprint verification and Wu et al. \cite{b7} detected both the principal lines and wrinkles for personal authentication. However, using limited line feature cannot achieve high recognition accuracy due to the similarity of principal lines among many subjects. Thus, much attentions have been directed towards the texture representation of palmprint image which carries rich discriminative direction features. Zhang et al. \cite{b8} designed an online palmprint identification system by utilizing a normalized 2-D Gabor filter to extract a special directional information of palmprint image, which plays a heuristic role in the later development of direction based methods. Since then, various direction based palmprint recognition methods were proposed. One of the most representative work is the competitive code method \cite{b9}, in which six Gabor filters with different directions were convolved with the palmprint image and the direction of the filter generating the strong filtering response was extracted as the dominant direction of palmprint image. Subsequently, several improved versions of competitive code methods were presented \cite{b10}\cite{b11}\cite{b12}. In addition, instead of extracting one direction feature, the binary orientation co-occurrence vector (BOCV) \cite{b13} binarized the filtering responses on six directions and Zhang et al. \cite{b14} extended the BOCV to E-BOCV by filtering out the fragile bits of the BOCV. Sun et al. \cite{b15} encoded the results of palmprint image convolving with three incorporated Gaussian filters on three directions.

Quite recently, the study of local texture representation is very active and various local texture descriptors are proposed. Among them, local binary pattern (LBP) \cite{b16} is an effective texture descriptor due to its efficiency and simplicity. Further, local direction pattern (LDP) \cite{b17} extended the LBP from image intensity space to gradient space for facial image representation achieving promising performance. Moreover, because of a plenty of line features of palmprint image, Luo et al. \cite{b18} proposed the local line direction pattern (LLDP) for palmprint recognition, in which the line detectors, such as Gabor filter and MFRAT \cite{b6}, were adopted to convolve with the palmprint image and the direction index with the maximum and minimum filtering responses were encoded.

It is seen that the aforementioned works are all based on the dominant direction features of palmprint image. Conventional popular methods generally extract only one of the most dominant direction feature depending only on the filtering response. However, plenty of points in palmprint images have multiple dominant directions due to a number of crossing lines in palmprint images and there is not work exploiting the representation the multiple dominant directions. It is seen that both LLDP \cite{b18} and DOC \cite{b10} extracted the double directions based on the most two strong filtering responses. Nevertheless, there is no explanation of employing the direction of the second strong filtering response. Also, the second largest filtering response is not corresponding to the secondary dominant direction of a palmprint image.

In this paper, we propose a novel local multiple directional pattern (LMDP) to represent the multiple dominant directions of palmprint image. LMDP can not only precisely indicate the number of dominant directions of palmprint image but also effectively reflect the position and confidence of each dominant direction. Extensive experiments are conducted on three palmprint databases to validate the effectiveness of LMDP.

The remainder of this paper is organized as follows. Section II briefly reviewes previous works. Section III elaborates the local multiple directional pattern of palmprint image and Section IV presents the experimental results and analysis. Finally, Section V concludes this paper.

\section{Previous works}

\subsection{LBP}
Local binary pattern (LBP) \cite{b16} is one of the most powerful texture descriptor with high robustness to rotation and illumination changes and low computational complexity. LBP operator labels every pixel in an image by thresholding its neighboring pixels against the center value. The LBP is scale invariant since the differences between the center pixel and the neighboring pixels are not affected by changes of mean luminance. To achieve rotation invariance, an improved LBP mode is generated by bit-wise right shift on the original LBP string and then the pattern with the unique minimum value is selected as the rotation invariant LBP. Moreover, some ``uniform" local binary patterns, which is the binary string that have at most two bitwise spatial transitions (bitwise 0/1 changes), account for the fundamental information of texture. A number is given to describe each of the ``uniform" patterns by summing up the number of ``1" in it, and the ``non-uniform" patterns are assigned to a special unique number. The uniformed LBP is recorded as LBP$_{P,R}^{riu2}$.

\subsection{Local directional pattern}
Inspired by LBP, local directional patterns (LDP) \cite{b17} is designed focusing on the image with rich line features. LDP is an eight-bit binary code calculated by comparing the edge responses of different directions in a local $3\times3$ neighborhood. Specifically, given a central pixel in an image, eight directional edge responses are calculated by convolving Kirsch edge masks with the pixel. The top $k$ responses are selected and corresponding directional bits are marked as 1 and the remaining bits are set to 0.

Extended by LDP, the enhanced local directional pattern (ELDP) \cite{b19} is generated by encoding the top-two edge responses as $t_1\times8+t_2$, where $t_1$ and $t_2$ are the directional index number of the first and second largest edge responses, respectively. Comparatively, local directional number (LDN) \cite{b20} pattern encodes the local directional information based on the maximum and minimum edge responses as $t_1\times8+t_8$, where $t_1$ and $t_8$ denote the directional index number of the maximum and minimum edge responses, respectively.

Luo et al. \cite{b16} further proposed a local line directional pattern (LLDP) to represent the palmprint image. Instead of using the gradient edge feature, LLDP uses the line directional feature space, which is obtained by convolving twelve line detectors, such as MFRAT and Gabor filters, with palmprint image. Based on these convolved results, similar encode schemes as LDP, EPLD and LDN are respectively adopted to produce different kinds of LLDP for palmprint images.

\section{Local multiple directional pattern}

\subsection{Direction feature extraction of palmprint image}

The common rule to extract the direction feature of palmprint images is using a set of line detectors with different directions to characterize the direction feature of palmprint images. The real part of Gabor filter is a powerful tool for direction feature extraction due to its line-like structure, which has the following general form:

\begin{equation}
\begin{aligned}
G(x,y,\theta,\mu,\sigma)=\frac{1}{2\pi{\sigma}^2} &\exp(-\frac{x^2+y^2}{2{\sigma}^2}) \\
&\cos(2\pi\mu(x\cos\theta+y\sin\theta)),
\end{aligned}
\end{equation}
where $\mu$ is the radial frequency in radians per unit length, $\sigma$ is the standard deviation of the elliptical Gaussian along the $x$ and $y$ axis, respectively, which are empirically set as $\mu$=0.11, and $\sigma$=5.6179. The ranges of $x$ and $y$ are the sizes of the filter and both of them are empirically set to 35. $\theta$ controls the direction of the Gabor function. A serial of directions $\theta_{j}=(j-1)\pi/N_{o},(j=1,2,...,N_{o})$ are usually used so as to define a bank of filters, where $N_{o}$ is the number of directions of filters and $j$ is seen as the index of a direction. To precisely extract the direction feature, $N_{o}$ is empirically set to 12 in this paper. In direction extraction, the filters are convolved with an input palmprint image $I$ by

\begin{equation}
r_{j}(x,y)=-G(\theta_{j})*I(x,y),
\end{equation}
where ``*" is convolution operator. It is pointed out that the filter is upside down because the dark lines correspond to small values in gray-level palmprint images. In general, the filter which has the most similar direction with the point can obtain the maximum filtering response, which is actually the minimum convolved result. Therefore, the direction of the filter generating the minimum convolved result with the palmprint image is treated as the dominant direction of palmprint image:

\begin{equation}
\theta(I(x,y))=arg\mathop{\min}_{\theta_{j}}r_{j}(x,y).
\end{equation}

\subsection{Local multiple directional pattern of palmprint image}

\begin{figure*}[htpb]
\centering
\subfigure[]{\includegraphics[width=2in]{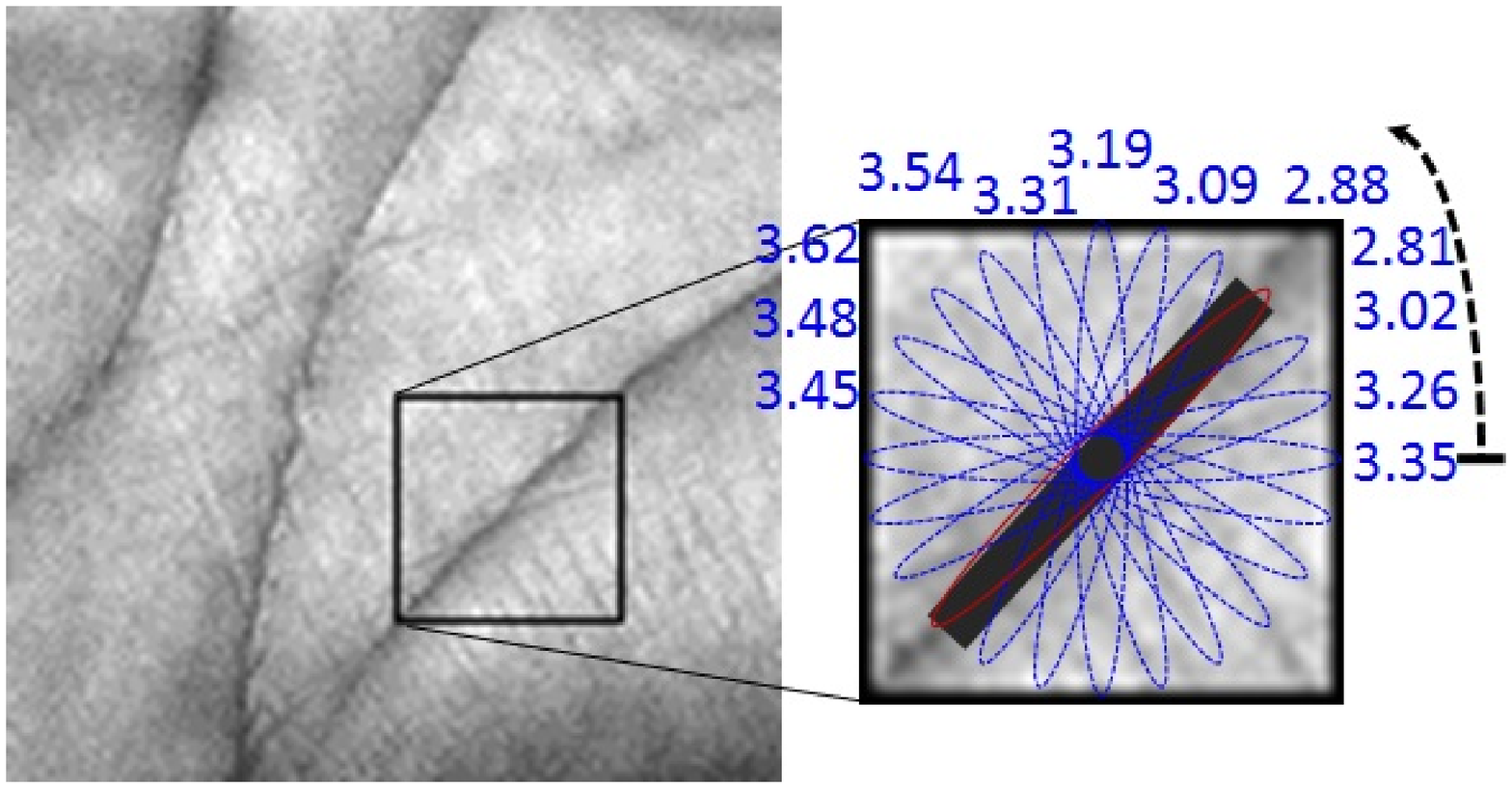}}
\subfigure[]{\includegraphics[width=1in]{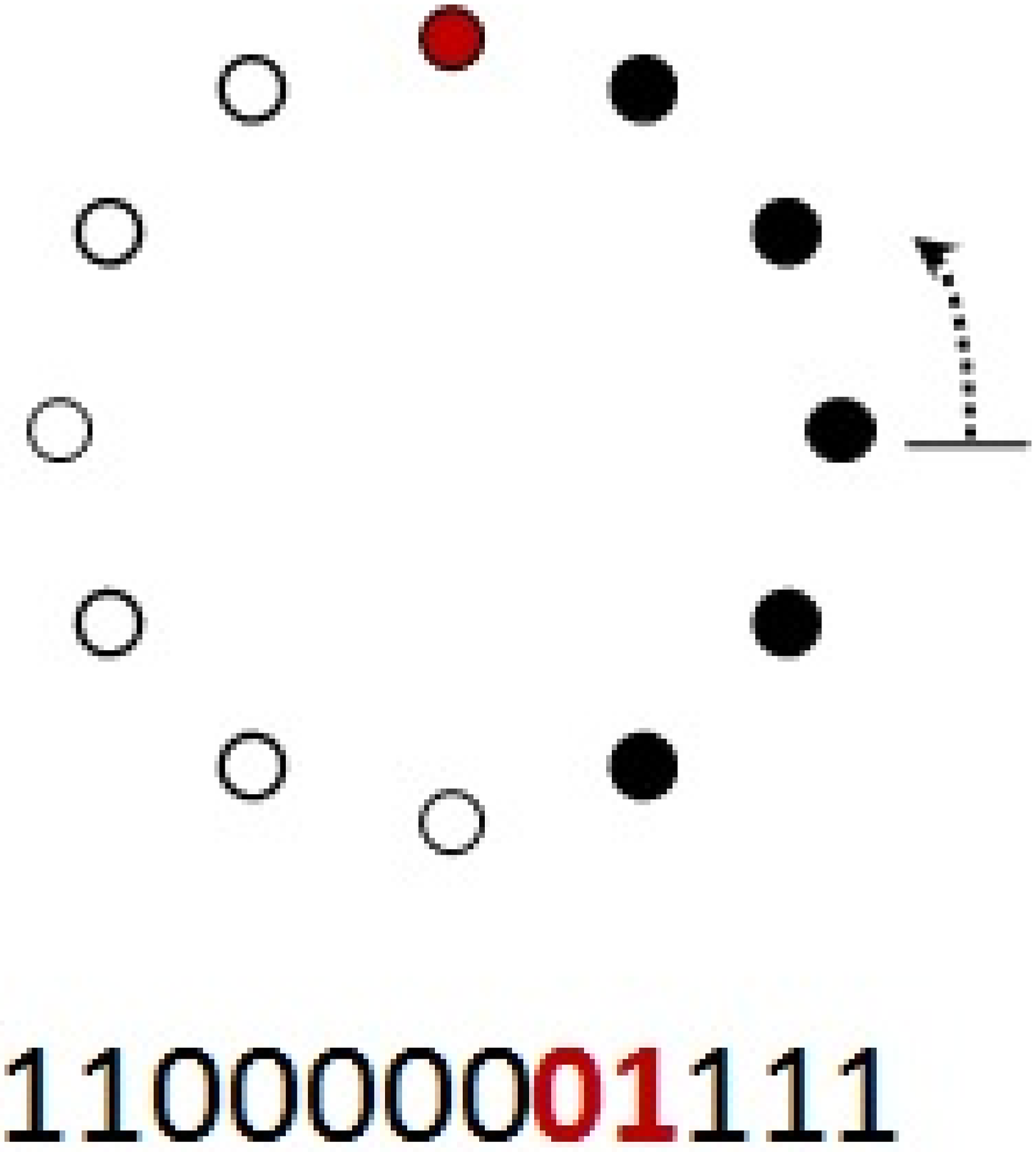}}
\subfigure[]{\includegraphics[width=2in]{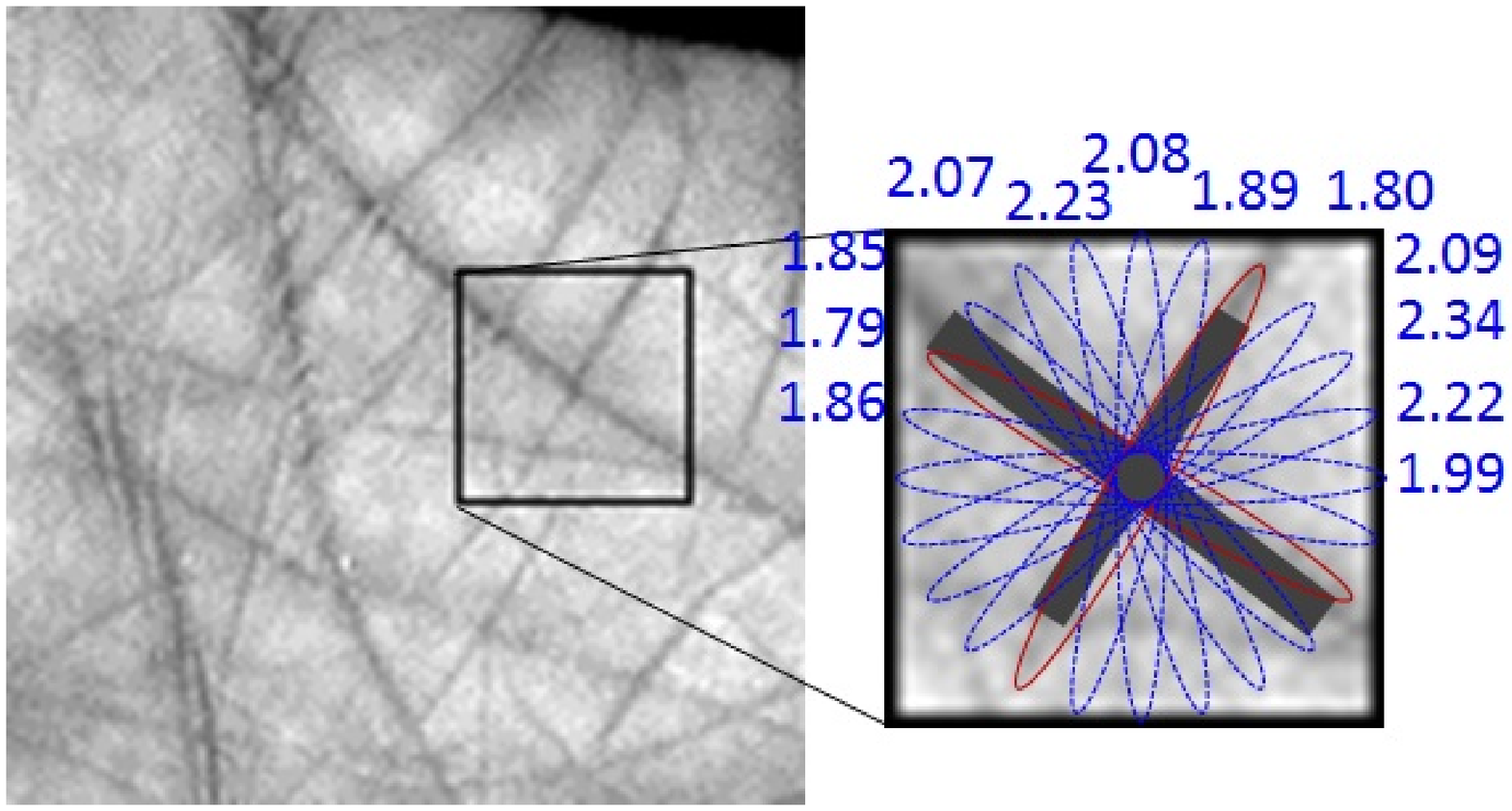}}
\subfigure[]{\includegraphics[width=1in]{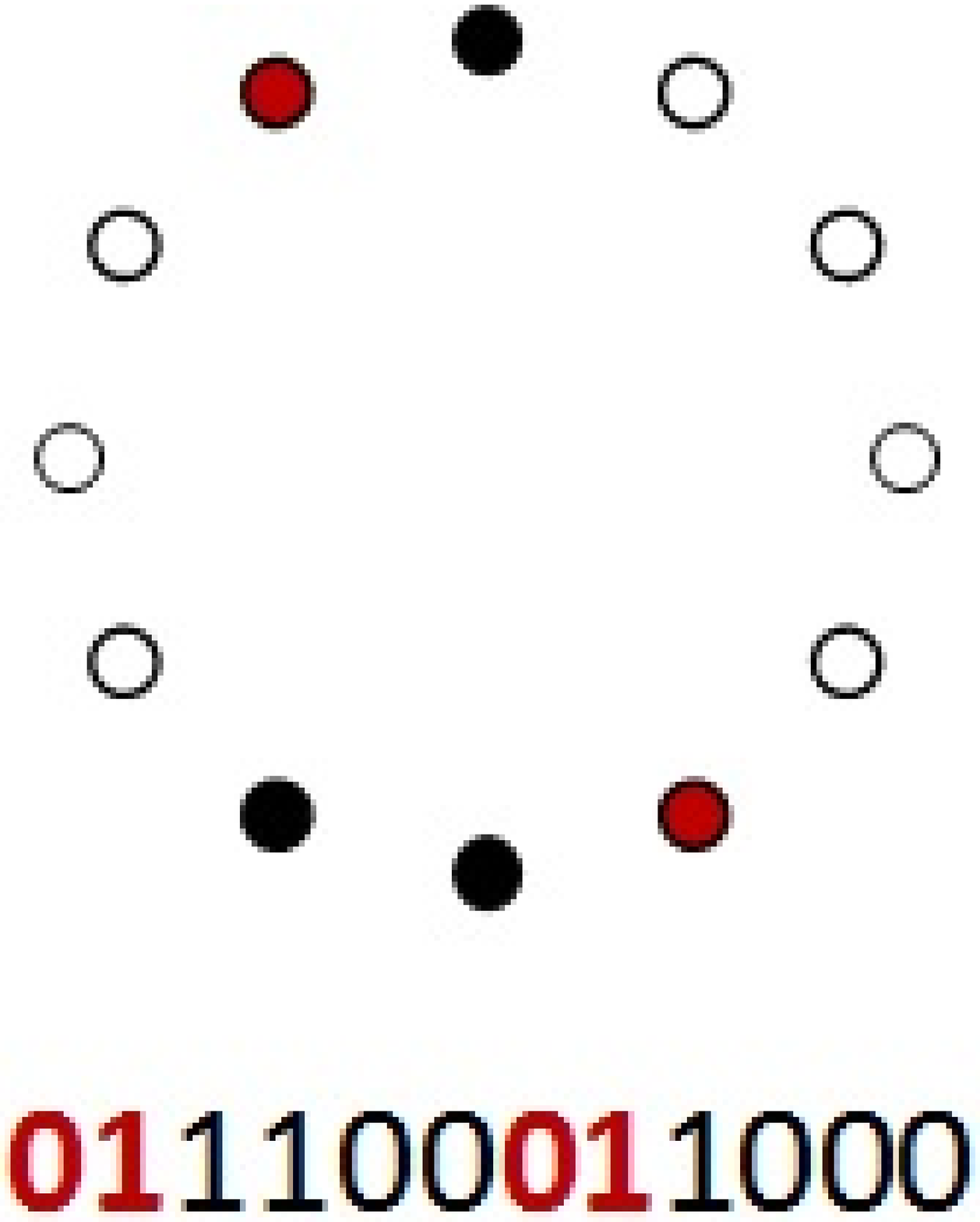}}
\caption{LMDP. (a) shows a point with one dominant direction and presents the convolved results on twelve directions and (b) depicts the LMBP of (a). Specially, the above circles demonstrate the circular property of LMBP, where black and white circles correspond to 1 and 0, respectively. The below 0/1 bit string is the LMBP of the point. In particular, arrow denotes the starting pattern and red represents the DP. (c) shows a point with double dominant directions and (d) presents the LMBP of (c).}
\label{Fig.lable.1}
\end{figure*}

In practice, palmprint image usually contains a number of crossing lines, which means that a number of points have multiple dominant directions. However, the basic rule of winner-take-all rule of direction extraction is based on the strongest filtering response selected from a bank of filters convolving with a palmprint image. In other words, only one of the most dominant direction can be extracted based on the winner-tale-all rule resulting to the loss of the other dominant direction features. So, conventional direction based methods cannot precisely characterize the multiple direction features of palmprint images.

The winner-take-all rule is based on the assumption that a filter which has the most similar direction with a point can produce the minimum convolved result. That is because the filter has maximum overlapped area with the line feature of palmprint image. Specifically, an upside down Gabor filter is line-like which has relatively small values on the ``line", and the dark lines of palmprint image also have relatively small scales. It is noted that the convolved result is directly proportional to the overlapping area between the line-like filter and the line feature. Thus, the filter which has the most similar direction with a line of a palmprint image can generate the minimum convolved result due to the maximum overlapping line area between them. Further, a closer direction of a Gabor filter to the dominant direction of the line can produce a larger overlapped area with the line so as to generate the smaller convolved result. In other words, between two neighbor directions of filters, a possible dominant direction of a point of a palmprint image should lie on side of the filter generating a smaller convolved value. A simple and effective way to represent the relationships between two filter responses on neighbor directions can be given as:

\begin{equation}
S=[s(r_1-r_{N_{o}}),s(r_2-r_1),...,s(r_j-r_{j-1}),...,s(r_{N_o}-r_{N_o-1})],\\
\end{equation}
where

\begin{equation}
s(x)=
\begin{cases}
1,& \text{x $<$ 0} \\
0,& \text{x $\geq$ 0}.
\end{cases}
\end{equation}

In other words, it is represented by ``1" if the convolved result on a direction is smaller than that on the adjacent clockwise direction, otherwise, it is represented as ``0". By assigning a binomial factor $2^j$ for each element $s(r_j-r_{j-1})$ in (4), it can be transformed into an unique local multiple direction pattern (LMDP) to characterize the dominant direction features of a palmprint image point.

\begin{equation}
LMDP=\sum_{j=1}^{N_o} s(r_j-r_{\varphi(j)})2^j,
\end{equation}
where $\varphi(j)$ denotes the adjacent clockwise direction index of $j$. It is noted that LMDP is circular and the direction indices of 1 and $N_o$ are adjacent.

\begin{equation}
\varphi(j)=
\begin{cases}
N_o,& \text{j=1} \\
j-1,& \text{2$\leq$j$<N_o$}.
\end{cases}
\end{equation}

LMDP can perfectly reflect the multiple directions of a point in a palmprint image. The pattern of ``01" in LMDP is named as Direction Pattern (DP) which essentially denotes a dominant direction, where ``1" means that the convolved result on the current direction is smaller than that on the clockwise neighbor direction, meanwhile ``0" denotes that it is also smaller than that on the counter-clockwise neighbor direction. The index of the ``1" in the DP represents the exact dominant direction and the number of DP is the number of dominant directions. Fig.1 shows the procedure of calculating LMBP, where the example pattern of ``1100000\textbf{01}111" represents that there is only one dominant direction at $3\pi/12$, and the ``\textbf{01}1100\textbf{01}1000" pattern denotes that the point of the palmprint image has two dominant directions at $4\pi/12$ and $10\pi/12$, respectively. Therefore, LMDP can not only denote the number of the dominant directions but also the exact position of each dominant direction.

\subsection{Coding LMDP}

It is easy to check that there is one-to-one correspondence between patterns of ``01" and ``10". So the DP number (DPN) of a LMDP can be obtained by:

\begin{equation}
DPN(LMDP)=\frac{1}{2}\sum_{j=1}^{N_o} |s(r_j-r_{\varphi(j)})-s(r_{\varphi(j)}-r_{\varphi({\varphi(j)})})|,
\end{equation}

The DPN correctly denotes the number of dominant directions of a point. The DP index (DPI), which is the index of ``1" of a DP in the LMDP, directly points out the position of a dominant direction. The DPI of a LMDP can be obtained as follows.

\begin{equation}
DPI(LMDP)=\{j|s(r_j-r_{\varphi(j)})-s(r_{\phi(j)}-r_j)=1\},
\end{equation}
where

\begin{equation}
\phi(j)=
\begin{cases}
j+1,& \text{1$\leq$j$<N_o$} \\
1,& \text{j$=N_o$}.
\end{cases}
\end{equation}

It is worth noting that different dominant directions have different confidences in the scenarios of DPN$\geq$2. In general, the confidence of a DP is determined by the filtering response. In practice, a filtering response is sensitive to small noise and rotation. To overcome the problem, we propose to use DP length (DPL) to determine the confidence of a DP, which is defined as the sum of the number of continuous ``1" on the right side and the number of continuous ``0" on the left side of the DP. Because the continuous ``1" and ``0" can effectively reflect the global dominance of the DP on the continuous area. The DPL of the DP in Fig.1(b) is 12, and the DPLs of DPs in Fig.1(d) are 5 and 7, respectively. It is seen that DPL can better represent the confidence of a DP and a larger DPL means a more stable dominant direction.

To effectively represent LMDP, we propose to use a label to represent LMDP. Specifically, after obtaining a LMDP, we calculate the DPN and DPI. The label of LMDP is directly set as the DPI when DPN=1. In the scenario of DPN=2, DPI is sorted in descending order first according to the DPL, and then to the filtering response in the case of the same DPL for two DPIs. After that, the LMDP is represented by a single label based on the primary DPI and secondary DPI. We have observed that very few pixels in palmprint images have more than two dominant directions, even less than 2\%. So a unique label is specially assigned to the LMDP with DPN$\geq$3. Thus, we designate the following operator for LMDP:

\begin{equation}
LMDP_{L}=
\begin{cases}
DPI,& \text{if DPN(LMDP)=1} \\
DPI_{1}\times{N_o}+DPI_{2},& \text{if DPN(LMDP)=2}\\
N_m, & \text{if DPN(LMDP)$\geq$3},
\end{cases}
\end{equation}
where $DPI_1$ and $DPI_2$ respectively are the primary and secondary DPIs, in the case of double dominant directions. Since the maximum value of $DPI_{1}\times{N_o}+DPI_{2}$ is $N_o^2+N_o-2$, $N_m=N_o^2+N_o-1$ is acceptable. As a result, a label of less than or equal to $N_o$ is the DPI of a single dominant direction, and a label of $N_m$ represents a point having more than two dominant directions. Additionally, other labels can exactly indicate the index number of the primary and secondary dominant directions.

\subsection{Block-wise LMDP descriptor}

To overcome the misalignments of palmprint images, we propose to use LMDP descriptor, a block-wise based histogram of LMDP$_L$, to represent palmprint images for palmprint recognition. Specifically, given a palmprint image, we first generate the LMDP$_L$ map $M$ of the palmprint image. Then, we uniformly divide $M$ into a set of non-overlapping small blocks $\{m_1,m_2,...,m_{N_b}\}$ with size of $p\times p$, where $N_b$ is the number of blocks. In general, a larger $p$ is suitable for more serious misalignment of images. In this paper, the block size is empirically set to 16$\times$16 pixels. For each $m_i$, we compute the histogram of LMDP$_L$ denoted by $h_i$ with the length of $N_m$, namely the maximum value of LMDP$_L$. After that, we concatenate all $h$s to create the global histogram $H$ of $M$ to form the LMDP descriptor of the palmprint image with the length of $N_bN_m$. Finally, we adopt the Chi-square distance to determine the similarity between two LMDP descriptors:

\begin{equation}
\chi^{2}(H^{A},H^{B})=\sum_{i=1}^{N_bN_m}\frac{(H_{i}^{A}-H_{i}^{B})^2}{H_{i}^{A}+H_{i}^{B}},
\end{equation}
where $H^A$ and $H^B$ represent two LMDP descriptors and $H_i$ is the value of $H$ at the $i-th$ bin.
\section{Experiments}
\subsection{Palmprint databases}

The PolyU database \cite{b21} contains 7,752 palmprint images collected from 386 palms of 193 individuals. The images were captured in two sessions with an interval of around two months. Each individual provides about 10 samples for both the left and right palms. It is noted that the 137th palm provided 17 images in the first session and the 150th palm provided only one image in the second session. As a result, a palm actually provides about 11 to 27 samples. The corresponding ROI images with size of 128$\times$128 pixels had been cropped.

The IITD database \cite{b22} consists of 2,300 contactless palmprint images collected from 460 palms corresponding to 230 subjects. 5 images were captured for each palm of a subject. The palmprint images of the IITD database have high intra-class differences due to user-pegs were not used for image acquisition. The corresponding ROI images with size of 150$\times$150 pixels had been extracted in the database.

The GPDS database \cite{b23} includes 1000 contactless palmprint images collected from the right palm of 100 volunteers, each of which provided 10 palmprint images. The palmprint images in the database are significant variant on rotations, translations and illuminations. The database also provides the corresponding ROI images, which are resized to 128$\times$128 pixels in the following experiments. Fig. 2 shows some typical palmprint ROI images selected from the PolyU, IITD and GPDS databases, respectively.

\begin{figure*}[htpb]
\centering
\subfigure[]{\includegraphics[width=0.8in]{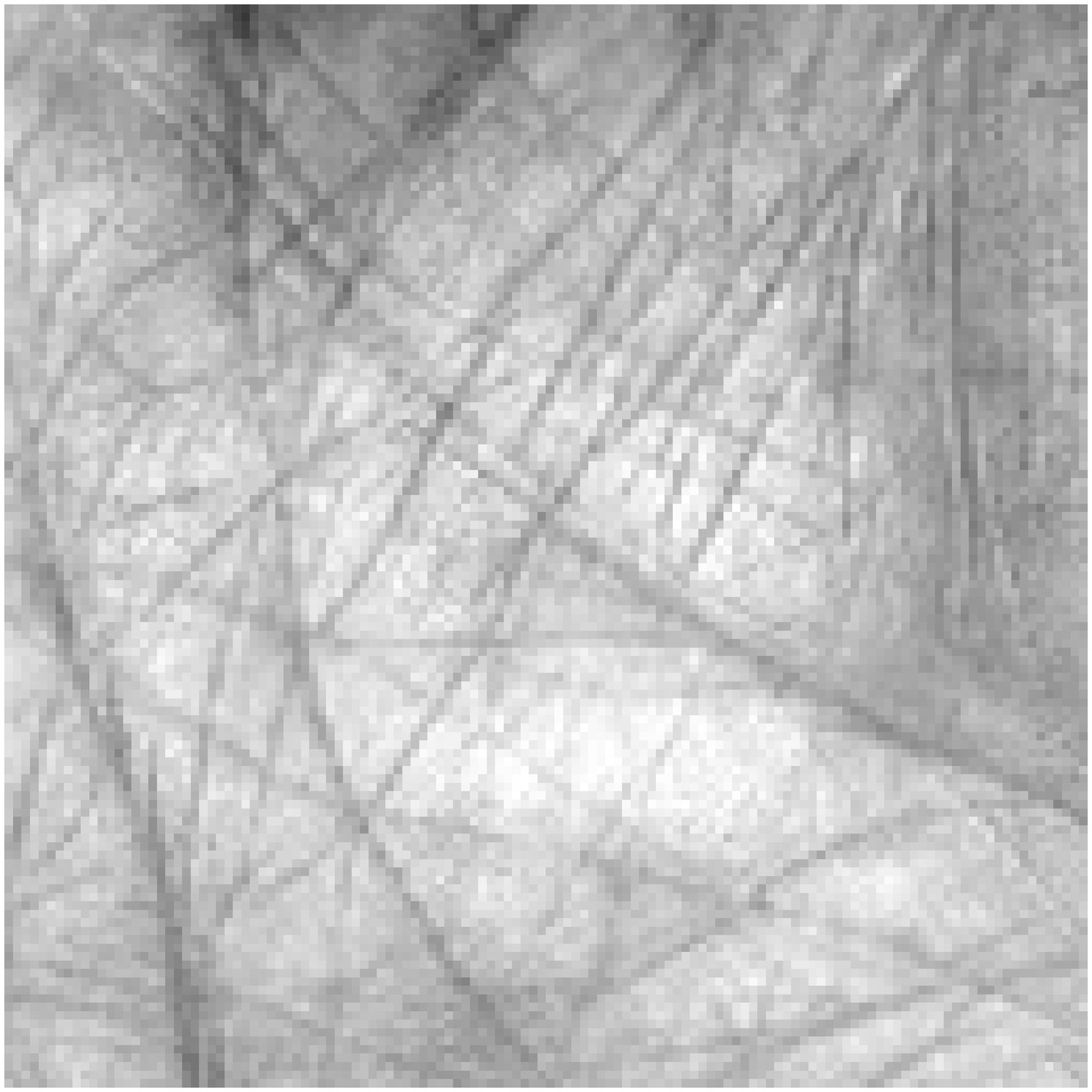}}
\subfigure[]{\includegraphics[width=0.8in]{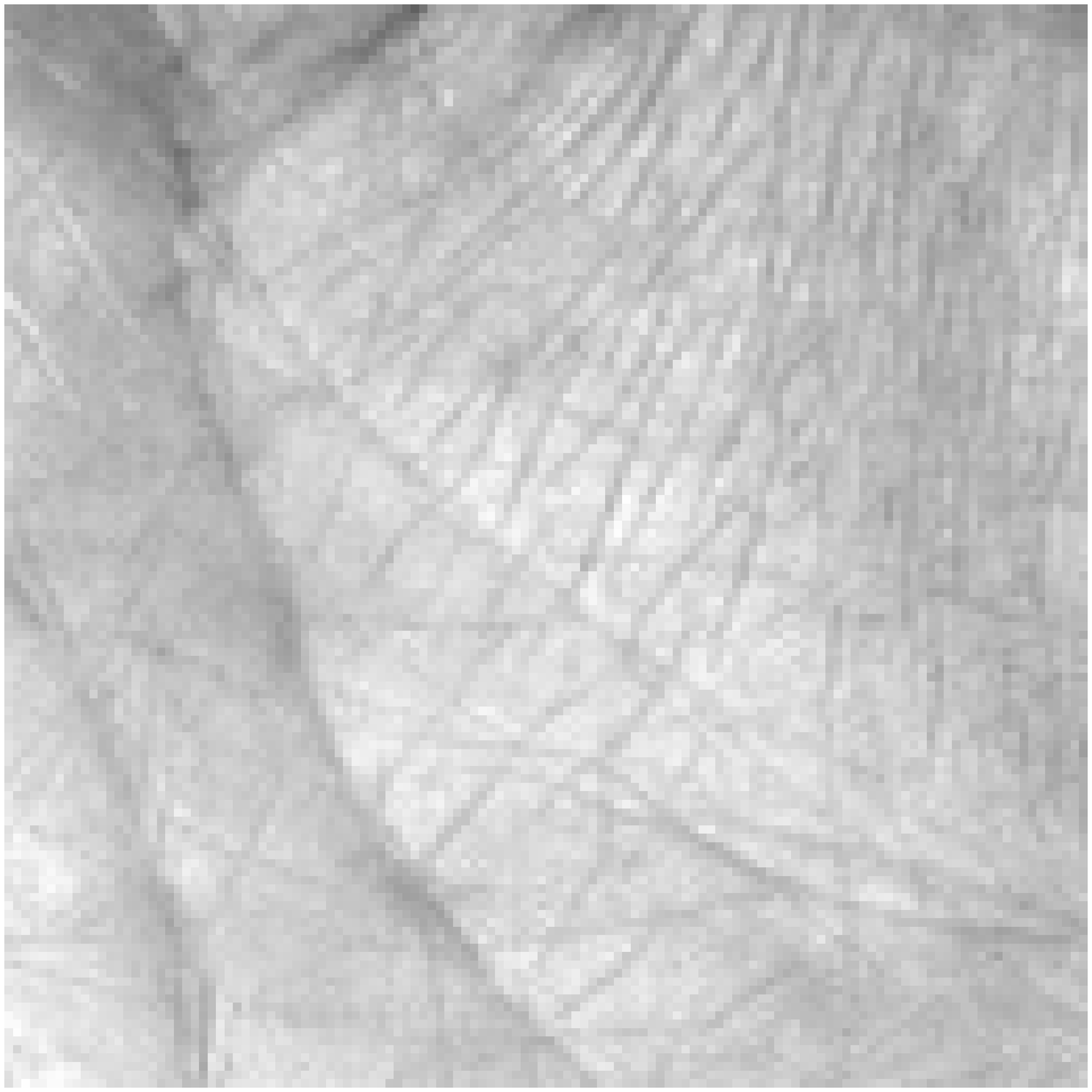}}
\subfigure[]{\includegraphics[width=0.8in]{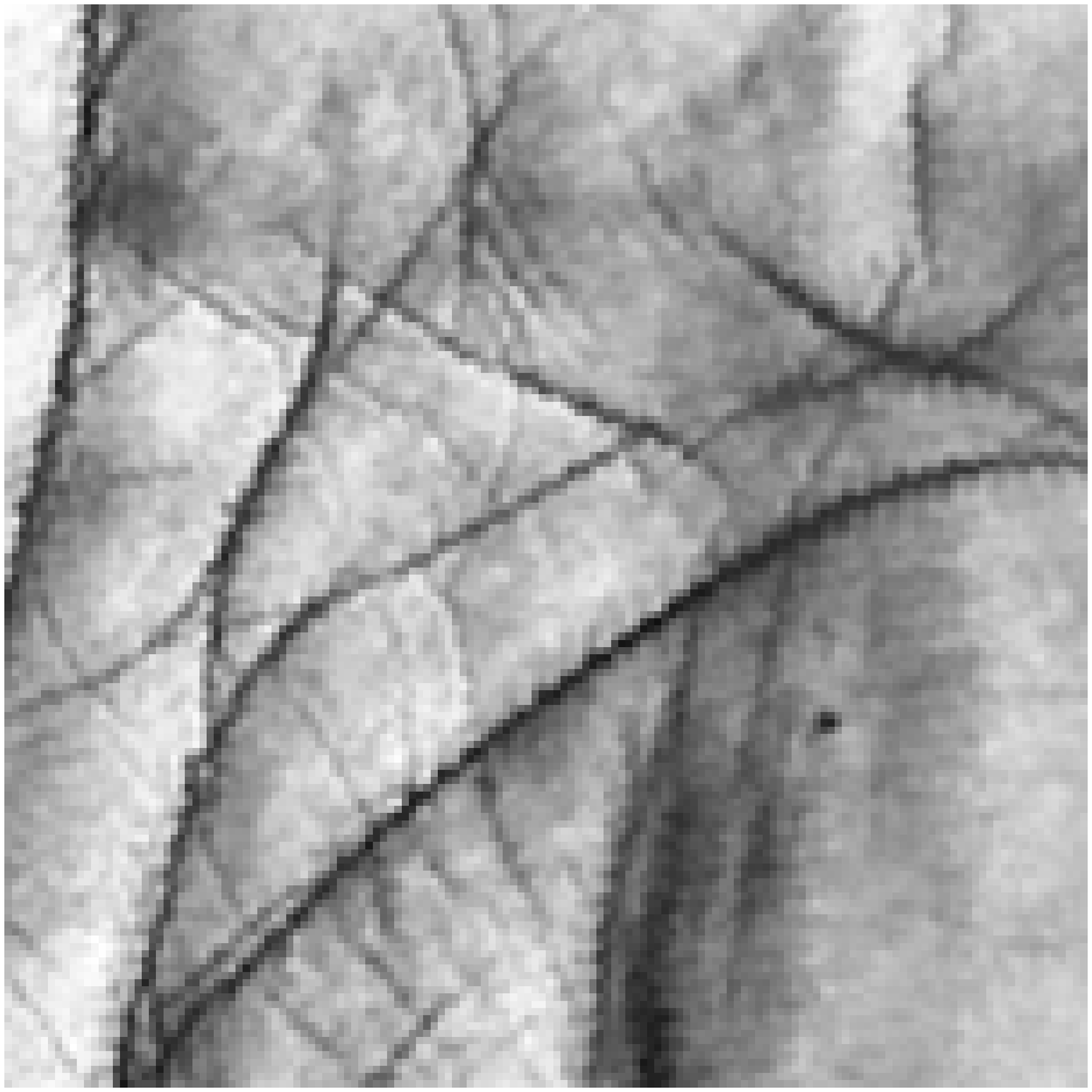}}
\subfigure[]{\includegraphics[width=0.8in]{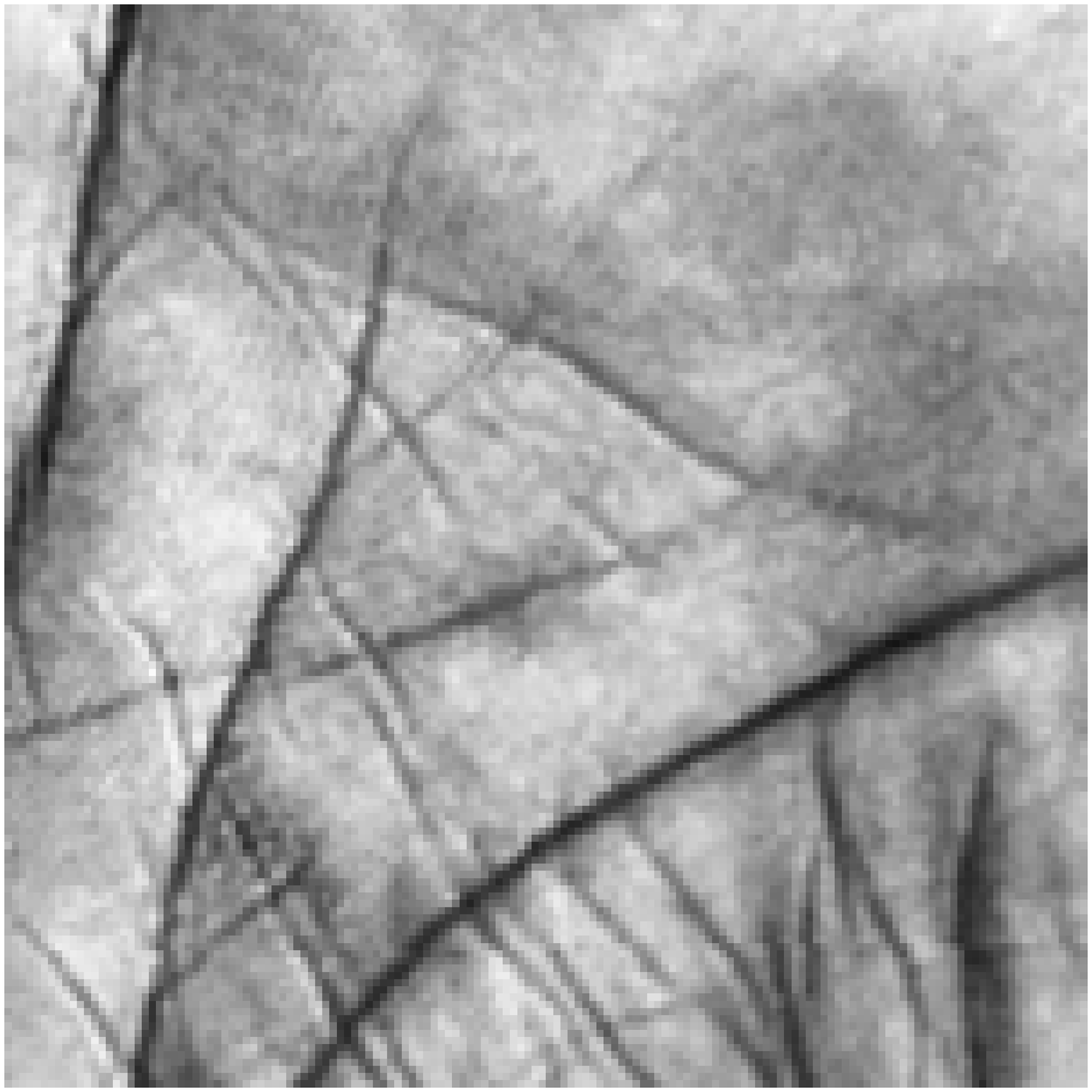}}
\subfigure[]{\includegraphics[width=0.8in]{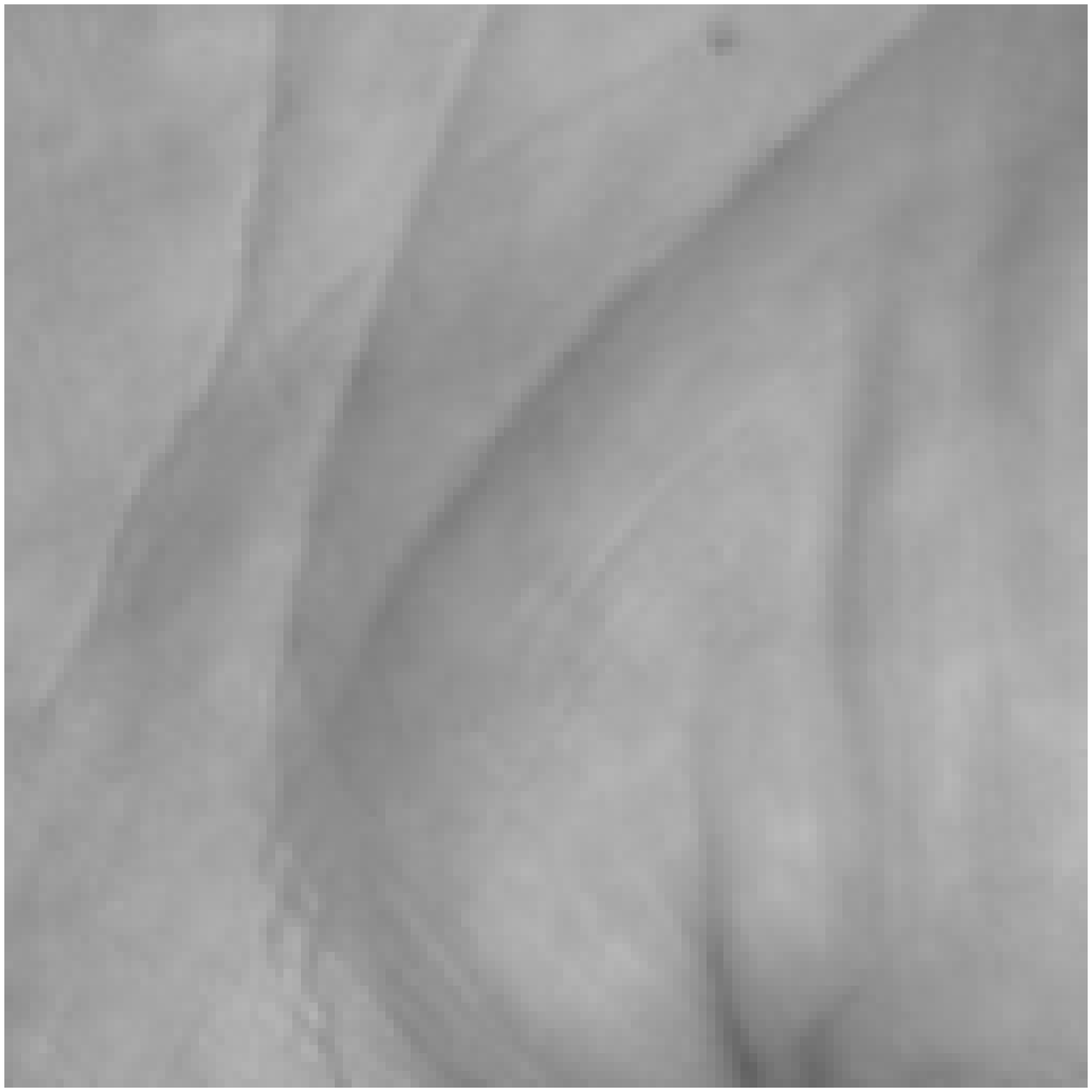}}
\subfigure[]{\includegraphics[width=0.8in]{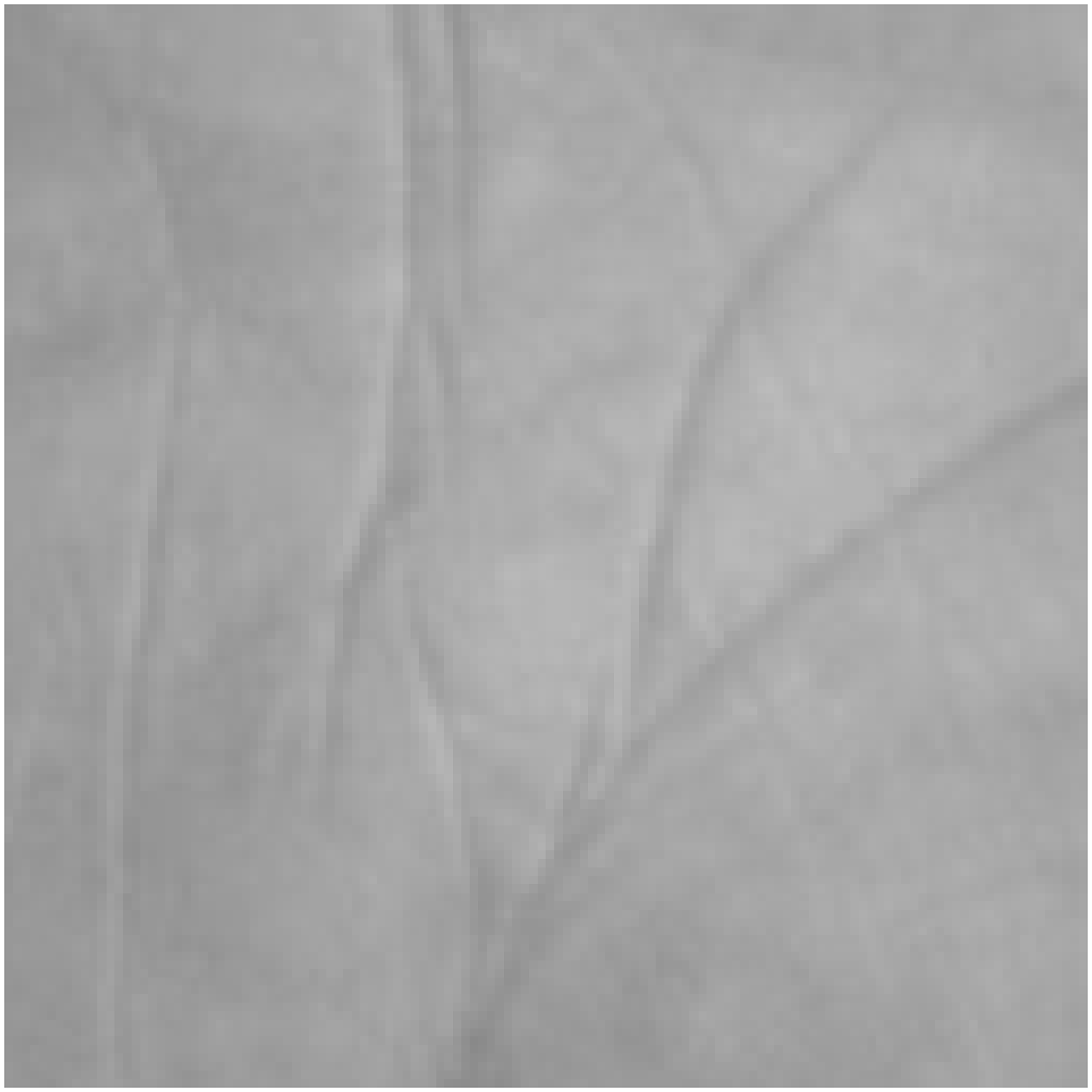}}
\caption{Some typical palmprint ROI images. (a)(b) are from the PolyU database; (c)(d) are from the IITD database and (e)(f) are from the GPDS database, respectively.}
\label{Fig.lable.2}
\end{figure*}

\begin{figure*}[htpb]
\centering
\subfigure[]{\includegraphics[width=1.6in]{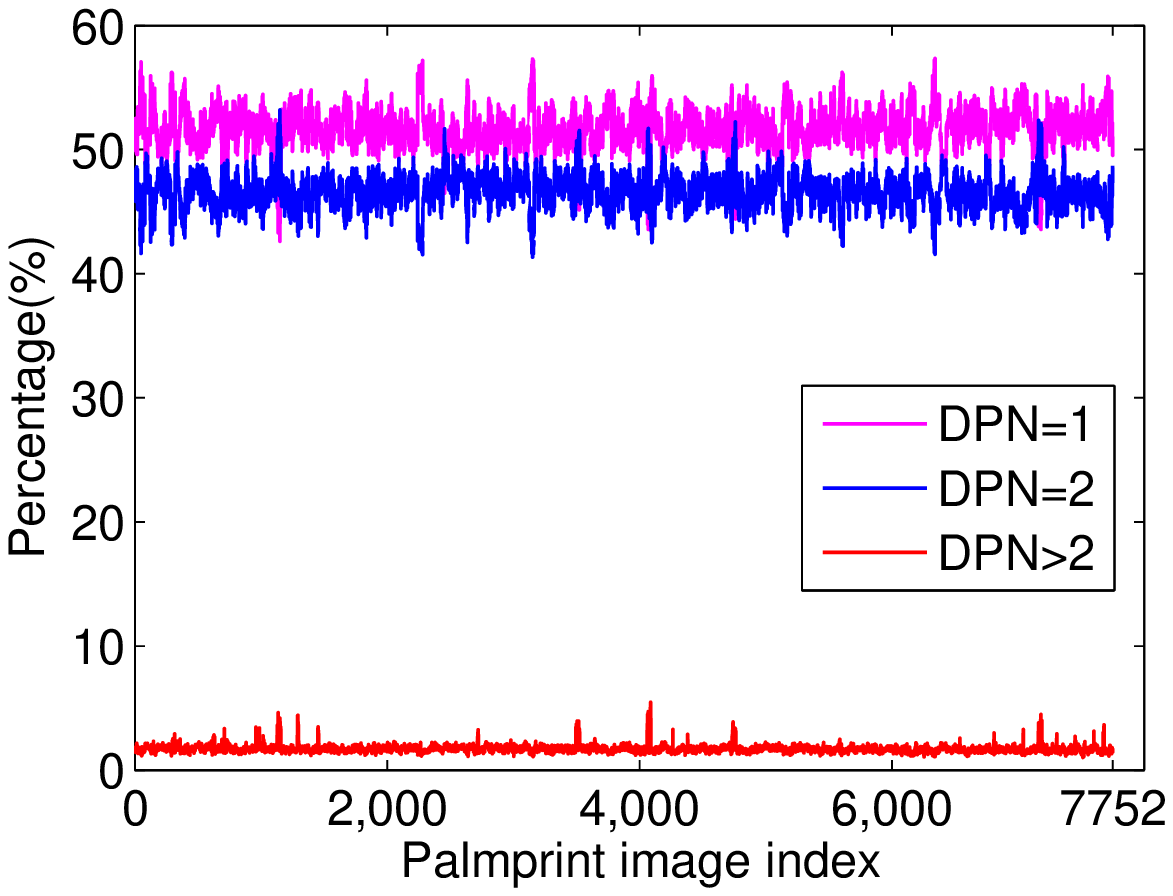}}
\subfigure[]{\includegraphics[width=1.6in]{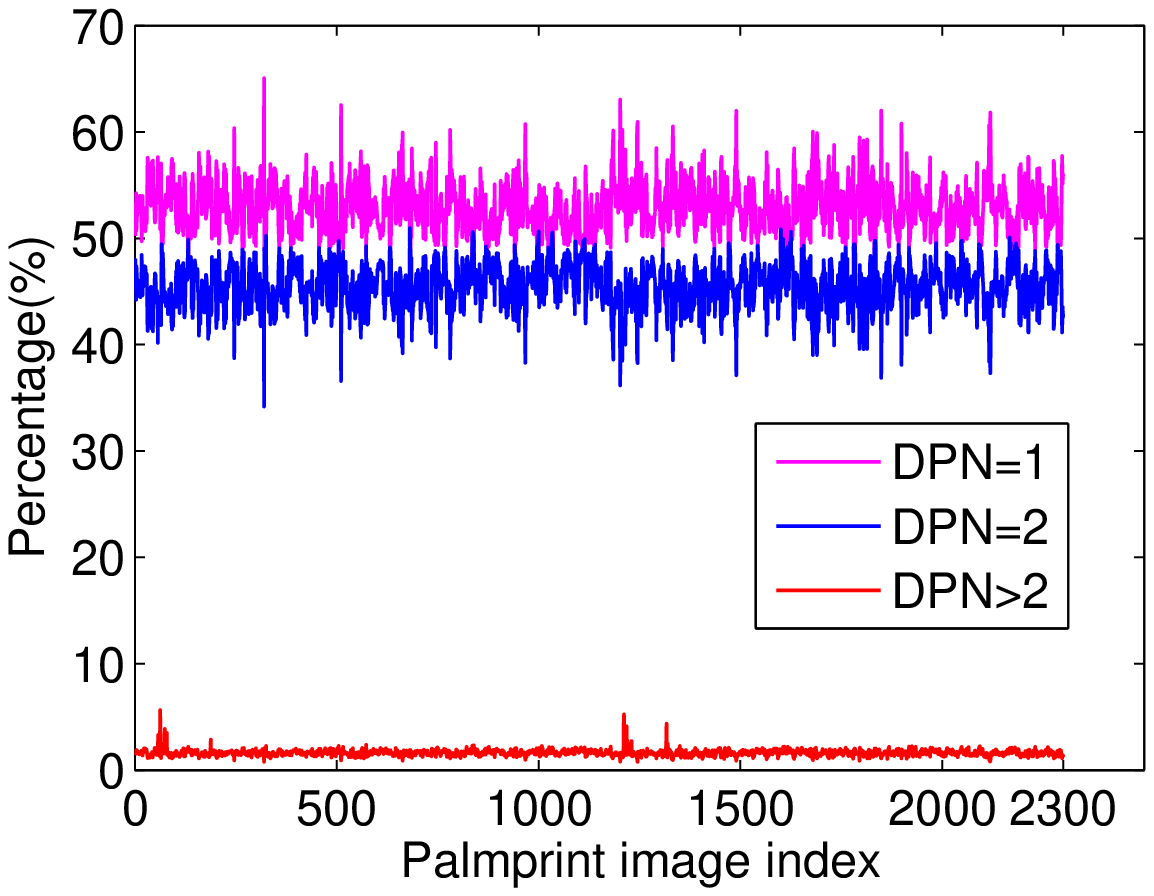}}
\subfigure[]{\includegraphics[width=1.6in]{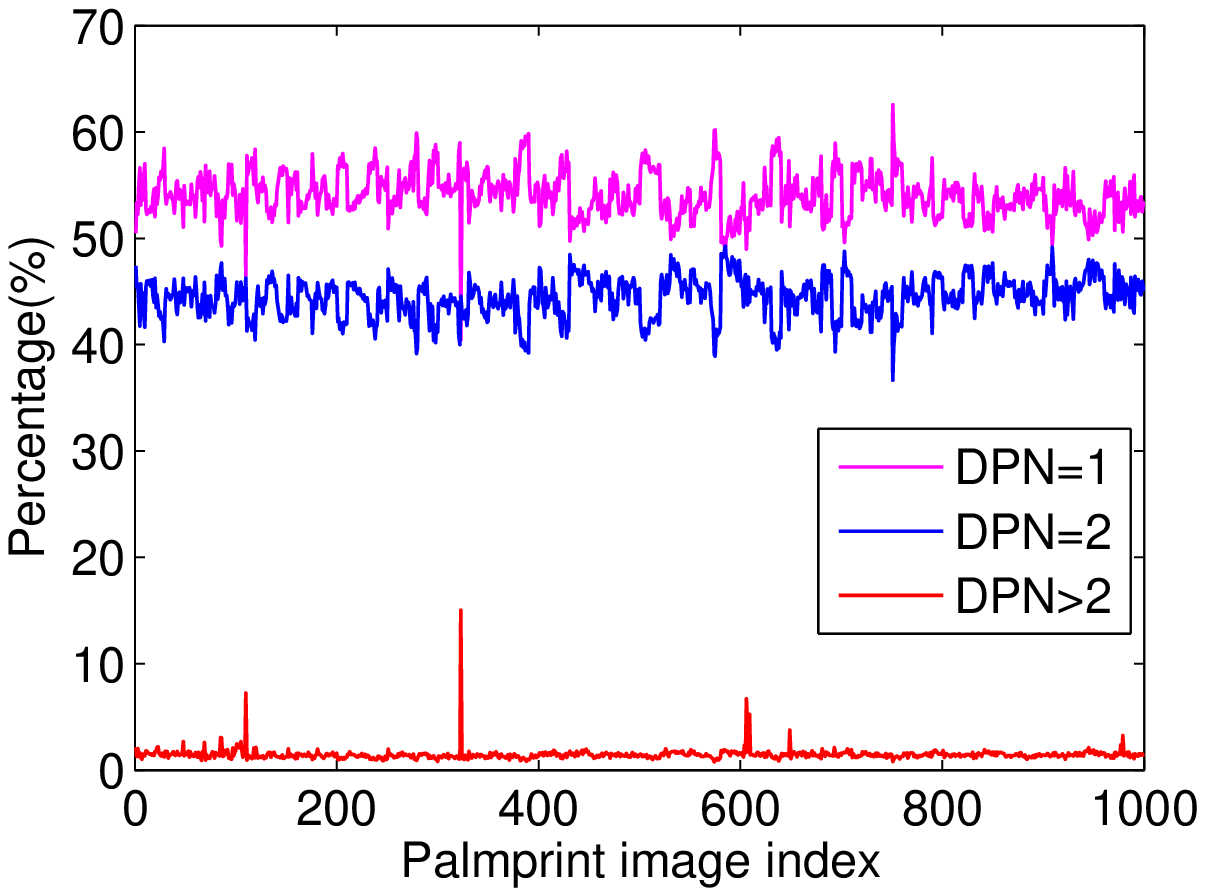}}
\caption{The DPN distribution of palmprint images. (a), (b) and (c) are DPN distributions on the PolyU, IITD and GPDS databases, respectively.}
\label{Fig.lable.2}
\end{figure*}

\subsection{Distributions of DPN}

To show DPN distribution of palmprint image, we calculate the DPN in each palmprint image. Specifically, given a palmprint image, we count the number of points with DPN=1, DPN=2 and DPN$\geq3$, respectively, so as to obtain the corresponding percentages of the three types of points. The distributions of DPN of palmprint images on three databases are depicted in Fig. 3, from which we can see that the points with DPN=1 and DPN=2 respectively are around 54\% and 45\% and points with DPN$\geq$3 are less than 2\% in most cases. In other words, there are a large number of points having double dominant directions and very few points have more than two dominant directions. Therefore, in addition to extracting the primary dominant direction, it is necessary to extract the secondary dominant direction to better represent the direction feature of palmprint. Meanwhile, since very few points have more than two dominant directions, it is reasonable to group the LMDPs with DPN$\geq$3 under a unique label.

\subsection{Experimental results}

We conduct both palmprint verification and palmprint identification experiments to evaluate the LMDP. Palmprint verification is a procedure of one-to-one comparison. A comparison is called as genuine matching if both palmprint images are from the same palm, otherwise the comparison is named as impostor matching. In the verification experiment of this study, each palmprint image in a database is matched with all other samples of the same database by using the proposed method to compute the incorrect genuine matching and incorrect impostor matching. After that, the equal error rate (EER) \cite{b8}, which is the point that the false acceptance rate (FAR) is equal to the false rejection rate (FRR), is adopted to estimate the performance of the proposed method.

Palmprint identification is a one-against-many matching process to determine the class label of a query sample. In the following identification experiments, for each database, we selected the first one, two and three palmprint images of each palm as training samples, and regarded the rest of images as the query samples. Under a query sample set, each query image will be compared with all templates in the corresponding training set. A query sample is classified into the class of the training sample that produces the smallest Chi-square distance with the query sample. That is, the nearest neighbor classifier is adopted in the palmprint identification. Finally, the rank-1 identification accuracy is calculated as the performance measure.

To demonstrate the effectiveness of the proposed method, we also test the performance of the conventional powerful descriptors, including LBP \cite{b16}, ELDP \cite{b19}, LDN \cite{b20} and LLDP \cite{b18}. It is noted that LLDP implemented four descriptors by respectively adopting the Gabor filter and MFRAT to produce filtering responses which are then respectively encoded using the EPLD and LDN scheme. Since both ELDP and LDN coding schemes in LLDP share similar performance, we implement both of them, that is, encoding the Gabor filtering results with EPLD and encoding the MFRAT filtering results with LDN, which are referred to LLDP\_Gab\_EPLD and LLDP\_MFRAT\_LDN, respectively. The local block-wise histogram, the size of which is similar as the proposed method, is established by using corresponding descriptor, and the Chi-square matching scheme is adopted. Moreover, previous state-of-the-art direction based coding methods, including the competitive code and ordinal code methods, are also implemented. In addition, SIFT features are usually extracted and fused with OLOF for contactless palmprint recognition \cite{b3}, namely SIFT\_OLOF, which is specially tested on the IITD and GPDS contactless palmprint databases. The experimental results on three databases are summarized in Table I-III, respectively, where the number under ``Identification accuracy" represents the corresponding training sample number per palm of palmprint identification. All algorithms are implemented under MATLAB 8.1.0.

\begin{table}[htpb]
\caption{Experimental results on the PolyU database (\%).}
\centering
\begin{tabular}{@{}lcccc@{}}\toprule
 & EER & \multicolumn{3}{c}{Identification accuracy} \\ \cmidrule(r){3-5}
 &     &   1  &   2   &   3   \\ \midrule
LBP$_{8,1}^{riu2}$   &  0.1235 &   62.67   &   69.74   &   74.05 \\
ELDP                 &  0.0488 &   89.90   &   93.15   &   94.13 \\
LDN                  &  0.0448 &   89.93   &   93.83   &   94.62 \\
LLDN\_Gab\_ELDP      &  0.0073 &   99.55   &   99.72   &   99.80 \\
LLDN\_MFRAT\_LDN     &  0.0105 &   99.26   &   99.46   &   99.62 \\
Competitive code     &  0.0261 &   95.16   &   96.99   &   97.86 \\
Ordinal code         &  0.0270 &   94.08   &   96.79   &   97.45 \\
LMDP                 &  \textbf{0.0059} &   \textbf{99.66} &   \textbf{99.80}   &   \textbf{99.82} \\
\bottomrule
\end{tabular}
\end{table}

\begin{table}[htpb]
\caption{Experimental results on the IITD database (\%).}
\centering
\begin{tabular}{@{}lcccc@{}}\toprule
 & EER & \multicolumn{3}{c}{Identification accuracy} \\ \cmidrule(r){3-5}
 &     &   1  &   2   &   3   \\ \midrule
LBP$_{8,1}^{riu2}$   &  0.1465 &   58.37 &   66.67   &   72.17 \\
ELDP                 &  0.0751 &   72.34 &   81.96   &   88.26 \\
LDN                  &  0.0725 &   73.37 &   82.32   &   88.70 \\
LLDN\_Gab\_ELDP      &  0.0337 &   86.47 &   92.75   &   95.43 \\
LLDN\_MFRAT\_LDN     &  0.0378 &   83.53 &   90.51   &   94.46 \\
Competitive code     &  0.0785 &   65.11 &   74.64   &   83.91 \\
Ordinal code         &  0.0744 &   61.14 &   72.54   &   82.17 \\
SIFT\_OLOF           &  0.0632 &   68.66 &   75.46   &   84.21 \\
LMDP                 &  \textbf{0.0264} &   \textbf{88.70} &   \textbf{93.91}   &   \textbf{96.20} \\
\bottomrule
\end{tabular}
\end{table}

\begin{table}[htpb]
\caption{Experimental results on the GPDS database (\%).}
\centering
\begin{tabular}{@{}lcccc@{}}\toprule
 & EER & \multicolumn{3}{c}{Identification accuracy} \\ \cmidrule(r){3-5}
 &     &   1  &   2   &   3   \\ \midrule
LBP$_{8,1}^{riu2}$   &  0.5280 &   48.11 &   60.25   &   72.57 \\
ELDP                 &  0.2929 &   67.56 &   81.87   &   91.57 \\
LDN                  &  0.3269 &   67.56 &   82.37   &   90.86 \\
LLDN\_Gab\_ELDP      &  0.2246 &   70.89 &   84.50   &   92.57 \\
LLDN\_MFRAT\_LDN     &  0.2198 &   73.44 &   86.38   &   93.29 \\
Competitive code     &  0.3961 &   51.56 &   64.25   &   76.43 \\
Ordinal code         &  0.4129 &   50.78 &   63.25   &   75.00 \\
SIFT\_OLOF           &  0.3563 &   56.82 &   68.78   &   78.47 \\
LMDP                 &  \textbf{0.1847} &   \textbf{76.67} &   \textbf{87.87}   &   \textbf{94.00} \\
\bottomrule
\end{tabular}
\end{table}

\subsection{Experimental results analysis}

Based on the experimental results, we can draw the following findings. First, LMDP, as well as LLDP, is based on the line direction space which are the most significant and discriminative features of palmprint image. Thus, LMDP achieves higher accuracy than the descriptors of LBP, ELDP and LDN.

Second, LMDP performs better than both LLDP and the direction based coding methods, including the competitive code and ordinal code methods. The main reason is that plenty points of palmprint images have multiple dominant directions, and LMDP can perfectly characterize the multiple dominant directions of palmprint images. Nevertheless, both LLDP and conventional coding methods just capture only one of the most dominant direction of palmprint images. In addition, LMDP employs more directions of filters than the coding based methods to relatively precisely extract the direction feature.

Third, it is notable that the LMDP descriptor performs much better than the direction based coding methods on IITD and GPDS databases. Because samples in both IITD and GPDS databases are contactless palmprint images, which have significant variances on rotations and translations. The block-wise statistic based features can effectively overcome the problems of rotation and translation changes.

\section{Conclusion}
Direction features of palmprint images with high discriminability have been successfully used for palmprint recognition. However, many points of palmprint images have multiple directions and conventional methods can only capture the most prominent one. In this paper, we propose a simple and novel local multiple direction pattern (LMDP) to effectively represent the multiple directions of palmprint images, of which both the position and confidence can be precisely indicated. A descriptor of block-wise LMDP is used as the feature vector of palmprint images in the matching stage. Extensive experimental results on three palmprint image databases demonstrate that LMDP is superior to the state-of-the-art direction based descriptors.

% conference papers do not normally have an appendix

% use section* for acknowledgment
\section*{Acknowledgment}
This paper is partially supported by the National Natural Science Foundation of China (Nos. 61300032, 61572284 and 61501230).

% trigger a \newpage just before the given reference
% number - used to balance the columns on the last page
% adjust value as needed - may need to be readjusted if
% the document is modified later
%\IEEEtriggeratref{8}
% The "triggered" command can be changed if desired:
%\IEEEtriggercmd{\enlargethispage{-5in}}

% references section

% can use a bibliography generated by BibTeX as a .bbl file
% BibTeX documentation can be easily obtained at:
% http://mirror.ctan.org/biblio/bibtex/contrib/doc/
% The IEEEtran BibTeX style support page is at:
% http://www.michaelshell.org/tex/ieeetran/bibtex/
%\bibliographystyle{IEEEtran}
% argument is your BibTeX string definitions and bibliography database(s)
%\bibliography{IEEEabrv,../bib/paper}
%
% <OR> manually copy in the resultant .bbl file
% set second argument of \begin to the number of references
% (used to reserve space for the reference number labels box)

% that's all folks
\end{document}